\documentclass[]{IAC_style}

\begin{document}

\IACpaperyear{2026}
\IACpapernumber{IAC-26,A3,IP,163,x107111}
\IAClocation{Antalya, T\"{u}rkiye}
\IACdate{5-9 October 2026}

\IACcopyrightA{}

\title{HyperLoop: Hyperspectral-Assisted Loop Closure Detection in 3D-SLAM for Planetary Exploration}

\IACauthor{Annie Bhalla$^{\orcidlink{0009-0000-2792-4551}}$}{1}{1}
\IACauthor{Dennis Rotondi$^{\orcidlink{0009-0008-7295-6729}}$}{1}{0}
\IACauthor{Riccardo Giubilato$^{\orcidlink{0000-0002-3161-3171}}$}{2}{0}
\IACauthor{Kai O. Arras$^{\orcidlink{0009-0001-3003-8857}}$}{1}{0}
\IACauthor{Wolfgang Stuerzl$^{\orcidlink{0000-0003-2440-5857}}$}{2}{0}

\IACauthoraffiliation{Institute for Artificial Intelligence (AI), University of Stuttgart,
  Universit{\"a}tsstra{\ss}e 32, 70569 Stuttgart, Germany.
  \normalfont{E-mail:~\authormail{anniebhalla16@gmail.com}}}
\IACauthoraffiliation{German Aerospace Center (DLR), Institute of Robotics and
  Mechatronics, 82234 Oberpfaffenhofen, Germany.}

\abstract{Autonomous navigation of planetary rovers remains a fundamental challenge in space
exploration, particularly in environments that are featureless, poorly illuminated,
or visually repetitive. Simultaneous Localization and Mapping (SLAM) systems based
on 3D Gaussian Splatting (3DGS) offer promising dense mapping capabilities; however,
their loop closure modules rely on RGB appearance descriptors that degrade under
perceptual aliasing--a common failure mode in extraterrestrial terrain where
physically distinct locations appear visually identical to optical sensors.
This paper presents HyperLoop, the first framework integrating hyperspectral imaging
(HSI) into the loop closure module of a 3DGS-based SLAM system.
Hyperspectral descriptors encode material-level spectral signatures that complement
RGB appearance features, providing a modality inherently robust to illumination
changes and visual repetitiveness.
Two HSI descriptor families are introduced: the Bag-of-Spectral-Words (BoSW)
model, which encodes spectral material distributions via TF-IDF weighted
frequency histograms, and the PCA-K VLAD (PCAK-VLAD) model, which captures
first-order residual statistics of PCA-compressed spectral signatures.
Both are evaluated in combination with SIFT-VLAD and NetVLAD RGB descriptors
under Strict-AND and Weighted decision-level fusion strategies.
The framework is assessed on a custom multi-modal dataset collected at the
Moon-Mars Outdoor Test Site (MMOTS) of the German Aerospace Center (DLR),
comprising synchronized RGB-D and hyperspectral sequences totalling 65.5~m
across four planetary analog traverses, with evaluation conducted on the
MMOTS-WS loop sequence (8~m).
A key empirical finding is that the sensor's Red-NIR spectral range (600--860~nm)
provides limited discriminability for planetary analog rock terrain: the best fusion configuration (SIFT-VLAD + BoSW, Strict-AND) achieves
5.04~cm aligned ATE RMSE, within 1.8\% of the best RGB-only configuration
(4.95~cm, SIFT-VLAD alone). The improvement over the NetVLAD baseline (8.81~cm) is attributable
entirely to replacing NetVLAD with SIFT-VLAD, not to HSI fusion;
the sensor's Red-NIR spectral range is identified as the primary
limiting factor for the HSI contribution.
This motivates future adoption of shortwave infrared (SWIR) sensors,
which capture diagnostic mineral absorption features absent from the
600--860~nm range, for planetary SLAM missions.
}
\IACkeywords{Planetary SLAM; Hyperspectral Imaging; Loop Closure Detection; 3D Gaussian Splatting; Place Recognition; Sensor Fusion}

\maketitle
\thispagestyle{fancy}

\raggedbottom
\setlength{\textfloatsep}{8pt plus 2pt minus 4pt}
\setlength{\intextsep}{6pt plus 2pt minus 2pt}
\setlength{\floatsep}{6pt plus 2pt minus 2pt}

\section*{Nomenclature}

\noindent
\begin{tabular}{ll}
$\mathbf{s}$       & Pixel spectral signature $\in\mathbb{R}^{15}$ \\
$\mathbf{z}_i$     & PCA-projected pixel spectrum $\in\mathbb{R}^{6}$ \\
$\mathbf{d}_i$     & Local SIFT descriptor $\in\mathbb{R}^{128}$ \\
$\mathbf{c}_k$     & $k$-th cluster centre (RGB or spectral codebook) \\
$a_i$              & Nearest-neighbour cluster assignment \\
$\mathbf{v}_k$     & VLAD residual vector for cluster $k$ \\
$K$                & Codebook / vocabulary size \\
$S(\cdot,\cdot)$   & Cosine similarity score \\
$\tau$             & Similarity threshold \\
$\mathcal{D}$      & Set of detected loop closures \\
$\mathcal{G}$      & Set of ground-truth loop closures \\
$\alpha$           & 3DGS Gaussian opacity \\
$\beta$            & Weighted fusion coefficient \\
\end{tabular}

\section*{Acronyms/Abbreviations}

\noindent
\begin{tabular}{ll}
3DGS   & 3D Gaussian Splatting \\
ROS    & Robot Operating System \\
ATE    & Absolute Trajectory Error \\
BoSW   & Bag-of-Spectral-Words \\
DLR    & German Aerospace Center \\
FOV    & Field of View \\
HSI    & Hyperspectral Imaging \\
MMOTS  & Moon-Mars Outdoor Test Site \\
NIR    & Near-Infrared \\
PCAK-VLAD & PCA-K Vector of Locally Aggregated Descriptors \\
PGO    & Pose Graph Optimisation \\
RMSE   & Root Mean Square Error \\
SIFT   & Scale-Invariant Feature Transform \\
SLAM   & Simultaneous Localization and Mapping \\
SWIR   & Shortwave Infrared \\
TF-IDF & Term Frequency--Inverse Document Frequency \\
VLAD   & Vector of Locally Aggregated Descriptors \\
\end{tabular}

\newpage
\section{Introduction}

Simultaneous Localization and Mapping (SLAM)---the joint estimation of a
robot's pose and a map of its surroundings from on-board sensor data---has
reached remarkable maturity over the past two decades~\cite{slam-handbook}.
Classical geometrically-grounded pipelines now achieve centimetre-level
accuracy in structured environments, while recent semantic extensions such as
3D scene graph representations~\cite{rotondi2026scenegraphs} further enrich
these maps with object-level and spatial reasoning. Yet this progress has
been concentrated in information-rich, feature-dense settings: urban streets,
indoor offices, and structured facilities where visual cues are abundant and
predictable. Mapping in \textit{extreme environments}---planetary surfaces
dominated by featureless regolith, visually repetitive craters, and
mineralogically diverse rocks---remains a frontier where classical SLAM
assumptions break down.

Autonomous navigation of planetary rovers must contend with exactly these
conditions. Rovers must navigate unknown, unstructured terrain autonomously,
performing in-situ scientific analysis while building consistent environmental
representations. Unlike the structured, feature-rich urban environments common
on Earth, extraterrestrial terrain is dominated by craters, regolith, rocks,
and dust that create a visually repetitive landscape. This causes
\textit{perceptual aliasing}: distinct locations appear nearly identical to
optical sensors. In SLAM, this directly impairs \textit{loop closure detection}:
the critical process of recognising a previously visited location to correct
accumulated odometric drift. Failed loop closures produce corrupted maps and
irrecoverable trajectory divergence, threatening mission success.

Despite visual similarity, planetary terrain is often mineralogically
diverse. Rocks of different composition, regolith with varying particle
distributions, and exposed bedrock exhibit distinct spectral signatures
in the infrared range, differences entirely invisible to conventional
RGB cameras. Hyperspectral imaging (HSI) captures tens to hundreds of
narrow spectral bands per pixel, encoding material composition and
chemical properties as a high-dimensional fingerprint. These signatures
reflect \textit{what a surface is made of} rather than how it appears,
and are inherently more robust to illumination changes.

Recent advances in 3DGS~\cite{kerbl2023} have transformed dense visual
SLAM by enabling efficient, photorealistic scene representations.
LoopSplat~\cite{loopsplat2025}, the first 3DGS SLAM system with explicit
loop closure, demonstrates compelling performance in structured indoor
environments using NetVLAD~\cite{netvlad2016} descriptors. However,
its reliance on RGB appearance limits applicability to the visually
uniform terrain of planetary environments.

This paper introduces \textbf{HyperLoop}, extending LoopSplat by
integrating hyperspectral sensing into its loop closure module.
The key contributions are:
\begin{enumerate}[leftmargin=1.5em, noitemsep, topsep=2pt]
  \item A 3DGS-SLAM architecture integrating HSI loop closure detection, the
        first such combination, to the best of our knowledge.
  \item Two HSI descriptor families: BoSW and PCAK-VLAD, adapted from
        visual retrieval literature to spectral signatures.
  \item Evaluation of Strict-AND and Weighted decision-level fusion
        strategies combining RGB and HSI descriptors.
  \item The first empirical analysis of hyperspectral loop closure on
        a planetary analog dataset, yielding an empirically motivated
        spectral range hypothesis for future mission hardware selection.
\end{enumerate}

The remainder of this paper is organised as follows:
Section~2 covers relevant background;
Section~3 reviews related work;
Section~4 describes the HyperLoop methodology;
Section~5 presents the experimental setup;
Section~6 analyses descriptor performance;
Section~7 reports trajectory and loop closure results;
Sections~8 and~9 discuss findings and conclude.

\section{Background}

\subsection{SLAM and Loop Closure}

SLAM jointly estimates a robot's pose trajectory and map from on-board
sensor data~\cite{slam-handbook}. Sensor noise causes \textit{odometric drift};
loop closure corrects it by recognising a previously visited location,
inserting a constraint, and triggering Pose Graph Optimisation (PGO)~\cite{Grisetti2010ATO}.
The pipeline comprises (1)~\textit{place recognition} via descriptor similarity
and (2)~\textit{geometric verification} via spatial consistency checks.
In planetary terrain, both stages are challenged: perceptual aliasing
degrades appearance-based retrieval, and surface planarity limits geometric filtering.

\subsection{3D Gaussian Splatting and LoopSplat}

3DGS~\cite{kerbl2023} represents a scene as a collection of anisotropic
3D Gaussian primitives, each parameterised by position $\boldsymbol{\mu}$,
covariance $\boldsymbol{\Sigma}$, opacity $\alpha$, and view-dependent
colour encoded via spherical harmonics (SH, a basis for view-dependent appearance).
Differentiable splatting projects each 3D Gaussian as a 2D Gaussian; overlapping
splats are composited via depth-sorted alpha-blending, enabling gradient-based
optimisation of all parameters.

LoopSplat~\cite{loopsplat2025} implements a real-time dense RGB-D SLAM
system on 3DGS with explicit loop closure. It maintains \textit{submaps}
(groups of keyframes) for computational tractability. Tracking minimises
photometric and geometric losses. Loop closure candidates are retrieved
via NetVLAD cosine similarity and verified by computing the geometric
overlap of Gaussian splats between candidate submaps. Confirmed loops
trigger submap registration in Gaussian space followed by PGO.

\subsection{Hyperspectral Imaging}

A hyperspectral image is a data cube $H \times W \times B$, where $B$
is the number of spectral bands. Each pixel stores a \textit{spectral
signature}: reflectance at $B$ wavelengths, encoding material
composition. Raw measurements are converted to reflectance via
calibration~\cite{lihsi2019}:
\begin{equation}
  r = \frac{p_{\mathrm{raw}} - p_{\mathrm{dark}}}
           {p_{\mathrm{white}} - p_{\mathrm{dark}}},
  \label{eq:reflectance}
\end{equation}
where $p_{\mathrm{white}}$ and $p_{\mathrm{dark}}$ are white-reference
and dark-reference frames respectively. This removes illumination
intensity, making spectral measurements comparable across acquisitions.
An additional spectral correction $\mathbf{v} = \mathbf{C} \cdot \mathbf{r} + \mathbf{T}$
corrects for spectral cross-talk between mosaic filter elements.

\subsection{Multi-Modal Sensor Fusion}

Multi-modal sensor fusion combines information from heterogeneous
sensors to improve robustness beyond what any single modality
can achieve~\cite{khaleghi2013}.
Three levels of fusion are commonly distinguished.
\textit{Early (data-level) fusion} merges raw sensor measurements
before any feature extraction, requiring tight spatial and temporal
alignment between modalities.
\textit{Feature-level fusion} concatenates or projects intermediate
representations into a shared embedding space, and is applicable
when modalities share a common field of view.
\textit{Decision-level (late) fusion} combines independent
per-modality decisions or scores, and is applicable even when
modalities have mismatched resolutions, fields of view,
or update rates--making it the most flexible strategy for
heterogeneous sensor pairs.

\section{Related Work}

\subsection{Visual SLAM in Unstructured Environments}
Classical visual SLAM systems including ORB-SLAM~\cite{orbslam2015}
and ORB-SLAM2~\cite{orbslam2017} achieve strong performance in structured,
feature-rich indoor and urban scenes; however, all RGB-based approaches
degrade severely in texture-poor terrain where distinctive appearance cues
are absent~\cite{altawil2024}. Outdoor and planetary analog environments
compound this by adding illumination variation, featureless regolith, and
large-scale perceptual aliasing--distinct locations that appear visually
identical. These conditions cause place recognition to fail systematically,
leading to missed loop closures and unbounded trajectory drift.
Recent work on 3DGS-based relocalization in planetary-analog
terrain~\cite{peribanez2026} independently confirms that RGB appearance
alone is insufficient for reliable localization under these conditions.
No existing RGB-only SLAM system reliably solves loop closure
on planetary analog terrain.

\subsection{3DGS-Based Dense SLAM}
Neural Radiance Fields (NeRF)~\cite{nerf2020} introduced volumetric scene
representations capable of photorealistic rendering but required
slow offline optimisation incompatible with real-time robotics.
3DGS~\cite{kerbl2023} replaced implicit networks with an explicit set of
anisotropic Gaussian primitives, enabling real-time rendering.
LoopSplat~\cite{loopsplat2025} built the first loop-closed 3DGS SLAM system
and serves as the direct baseline for this work.
Concurrent work by Perib\'{a}\~{n}ez~et~al.~\cite{peribanez2026} demonstrates
3DGS-based visual relocalization on planetary-analog terrain and confirms that
standard 3DGS representations suffer from aliasing and sparse-view ambiguity
under challenging outdoor conditions--precisely the failure mode HyperLoop
targets at the loop closure stage.
None of these systems incorporate non-visual sensing into their
loop closure modules.

\subsection{Loop Closure Detection}
Classical place recognition uses visual vocabulary retrieval with hand-crafted
features; a comprehensive overview is provided by~\cite{arshad2021}.
NetVLAD~\cite{netvlad2016} improved robustness to viewpoint and illumination
by training on large-scale urban datasets, but this introduces a domain gap
for natural unstructured terrain where object-centric features are absent.
VLAD aggregation~\cite{jegou2010,netvlad2016} underpins both NetVLAD and
the hand-crafted descriptor baselines evaluated in this work.
To the best of our knowledge, all existing loop closure detectors in
3DGS-based dense SLAM operate on RGB appearance alone; under perceptual
aliasing, appearance similarity provides no reliable signal. LiDAR-based
and multimodal loop closure methods are well established in other SLAM
frameworks~\cite{arshad2021}, but none have been integrated into the
3DGS submap pipeline.

\subsection{Hyperspectral Imaging in Robotics and Remote Sensing}
HSI has been extensively used in remote sensing for material
classification and mineral mapping~\cite{lihsi2019}.
Orbital instruments such as CRISM on the Mars Reconnaissance
Orbiter~\cite{murchie2007} have successfully identified mineralogical
diversity on planetary surfaces, and drone-based surveys have demonstrated
discriminative rock spectral analysis at ground level~\cite{lee2021}.
Despite this maturity in remote sensing, integration of HSI into real-time
robotic SLAM systems remains entirely unexplored.
HyperLoop is, to the best of our knowledge, the first system to
integrate hyperspectral sensing into the loop closure module of a
3DGS-based dense SLAM pipeline, closing the gap between orbital
spectral mapping and ground-level autonomous navigation.

\section{Methodology}
\subsection{System Overview}

\begin{figure*}[t]
  \centering
  \includegraphics[width=\textwidth]{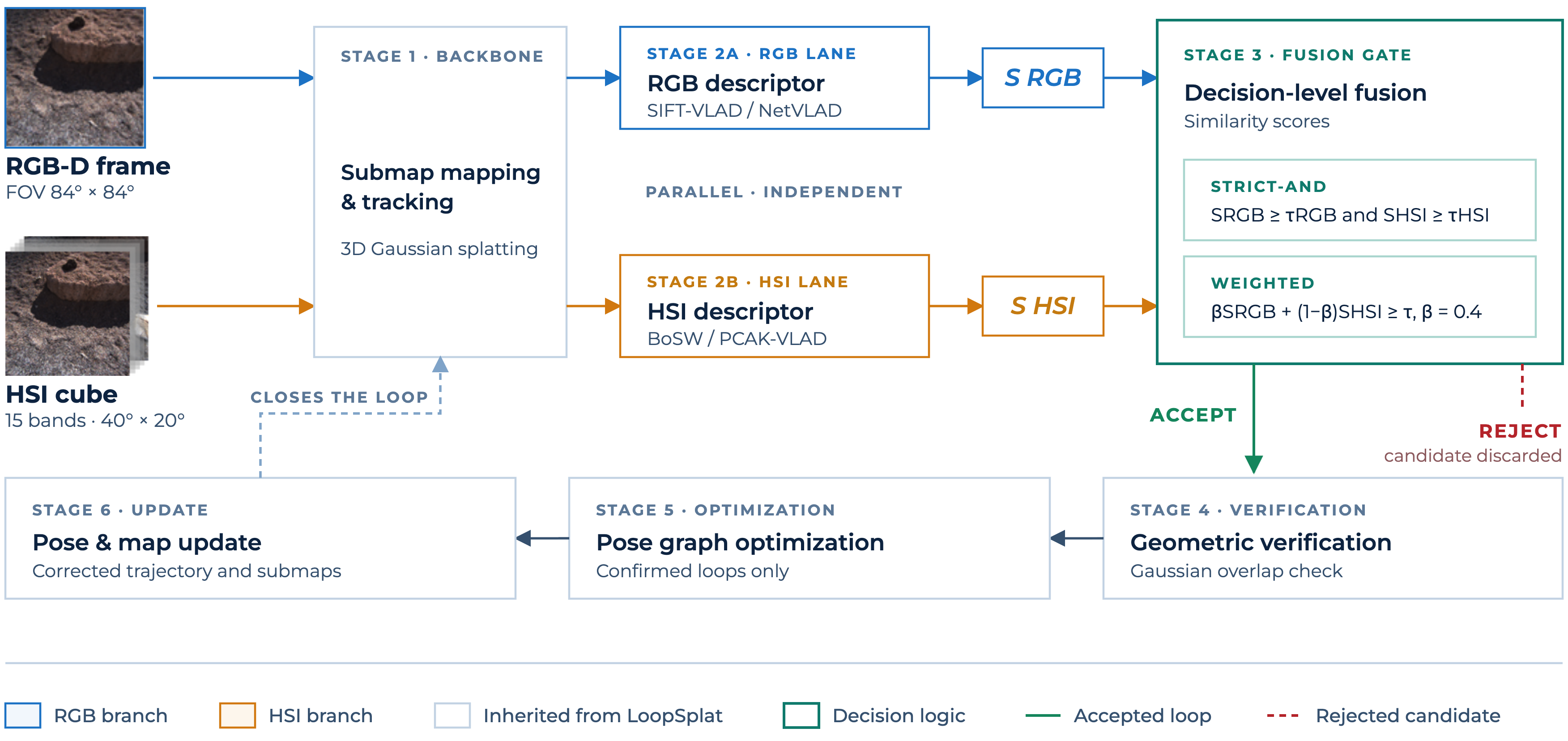}
  \caption{Overview of the HyperLoop framework. RGB and HSI streams
  are processed independently to yield complementary descriptors,
  which are combined via decision-level fusion before geometric
  verification and Pose Graph Optimization.}
  \label{fig:pipeline}
\end{figure*}

HyperLoop extends LoopSplat by augmenting its loop closure module
with a parallel hyperspectral processing pipeline, as illustrated
in Fig.~\ref{fig:pipeline}.

For each incoming RGB-D frame, the system performs tracking and
submap-based 3DGS mapping following LoopSplat. Simultaneously, for
each keyframe, two descriptor sets are computed: RGB descriptors from
the colour image and HSI descriptors from the temporally aligned
hyperspectral frame. When a new submap is finalised, loop closure
candidates are evaluated by computing pairwise similarity under both
modalities. A candidate is accepted only if it passes a decision-level
fusion gate followed by LoopSplat's geometric consistency check.
Confirmed closures are integrated into PGO to correct the trajectory.

\subsection{Outdoor Configuration}

The default LoopSplat configuration targets indoor RGB-D sequences.
Applied directly to MMOTS outdoor data, it produces catastrophic
trajectory divergence (ATE RMSE: 3380.89~cm). Two adaptations are
required. First, the \texttt{odometer} tracking mode is adopted,
which aligns consecutive RGB-D frames to produce accurate relative
pose estimates, replacing the constant-velocity model suited to
constrained indoor motion. Second, an inverse-depth L1 loss term is enabled to improve tracking
stability. When visual gradients are absent--as is common
on uniform regolith or gravel--a depth loss provides a dense geometric
constraint that anchors the pose estimate and prevents scale drift.
The inverse-depth formulation prevents distant points (sky, far terrain)
from dominating the loss due to their large absolute depth values,
improving convergence in outdoor scenes with large depth variation.
L1 loss is preferred over L2 loss as its constant gradient magnitude (independent of residual size) is more
robust to depth outliers from sensor noise, avoiding the
over-penalisation that would otherwise destabilise tracking.

\subsection{RGB Descriptor Extraction: SIFT-VLAD}

\begin{figure}[t]
  \centering
  \includegraphics[width=\columnwidth]{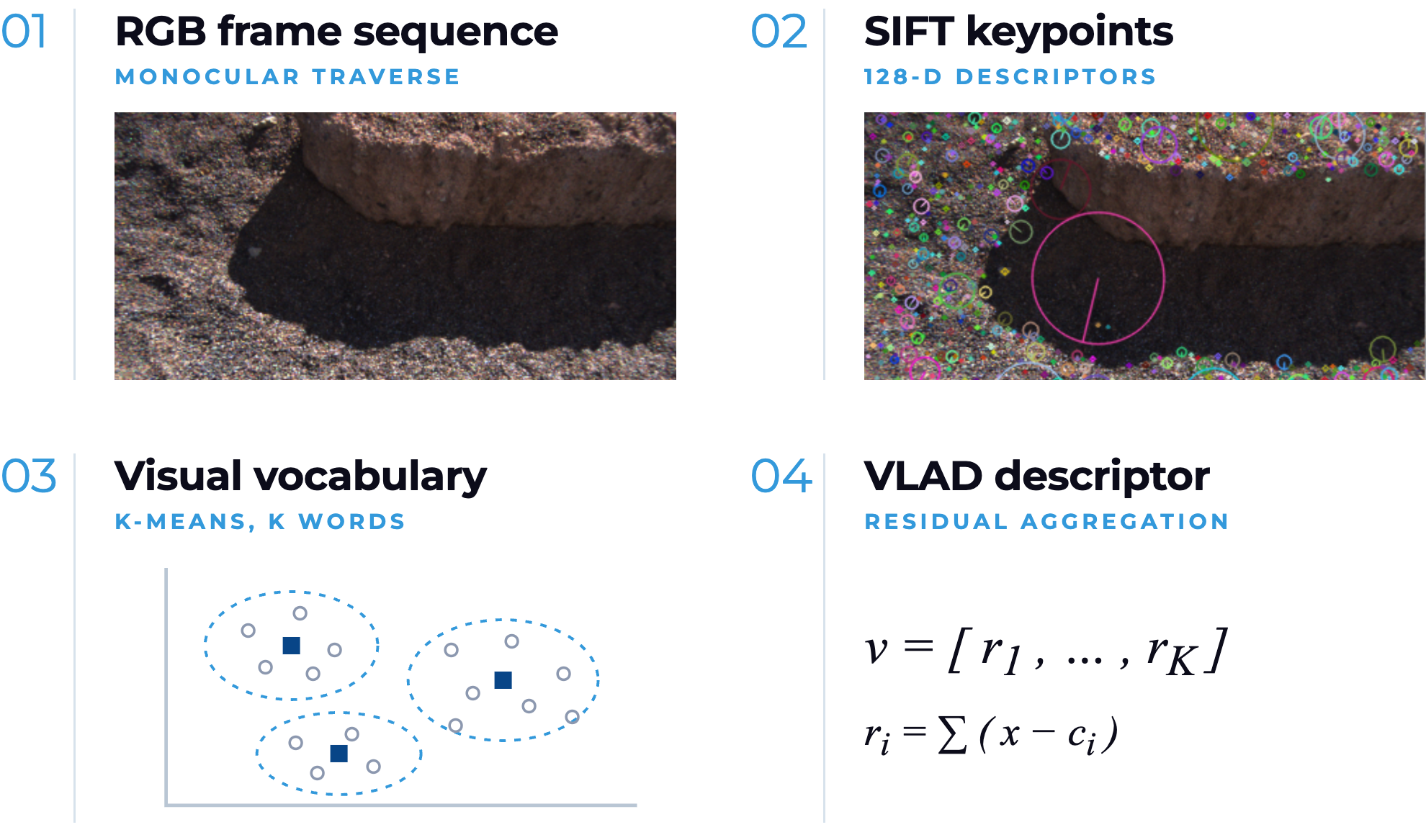}
  \caption{The SIFT-VLAD pipeline. SIFT keypoints provide
  128-dimensional local descriptors; K-Means clustering builds a visual
  codebook; VLAD aggregates residuals per cluster into a global
  descriptor invariant to scale, rotation, and illumination.}
  \label{fig:siftvlad}
\end{figure}

Two RGB descriptor models are evaluated. The first is NetVLAD~\cite{netvlad2016},
a VGG-16 (standard 16-layer CNN) based model with a trainable VLAD pooling layer producing
4096-dimensional descriptors, originally used in LoopSplat. The Pittsburgh30K
pre-trained checkpoint (VGG16-NetVLAD-Pitts30K) is used throughout, with random seed fixed to 0.
The second, and primary proposed contribution for outdoor terrain, is
SIFT-VLAD, illustrated in Fig.~\ref{fig:siftvlad}.

SIFT keypoints are detected across all input RGB frames, yielding
128-dimensional local descriptors. A visual codebook of $K$ cluster
centres is learned via K-Means on the training descriptor set.
The VLAD encoding aggregates descriptors by computing, for each
cluster $k$, the sum of residuals~\cite{jegou2010}:
\begin{equation}
  \mathbf{v}_k = \sum_{i:\,a_i = k}\!\left(\mathbf{d}_i - \mathbf{c}_k\right) \in \mathbb{R}^{128},
  \label{eq:vlad}
\end{equation}

where $a_i = \arg\min_k\|\mathbf{d}_i - \mathbf{c}_k\|$ is the
nearest-neighbour assignment. The final descriptor is formed by
concatenating all $\mathbf{v}_k$, flattening to $\mathbb{R}^{128K}$,
and applying L2 normalisation. Codebook sizes $K\in\{8,16,32,64\}$
are evaluated; $K=8$ yields the best separation on the MMOTS-WS
dataset (Section~6.1) and is adopted throughout.

\subsection{HSI Descriptor Extraction}

\begin{figure}[t]
  \centering
  \includegraphics[width=\columnwidth]{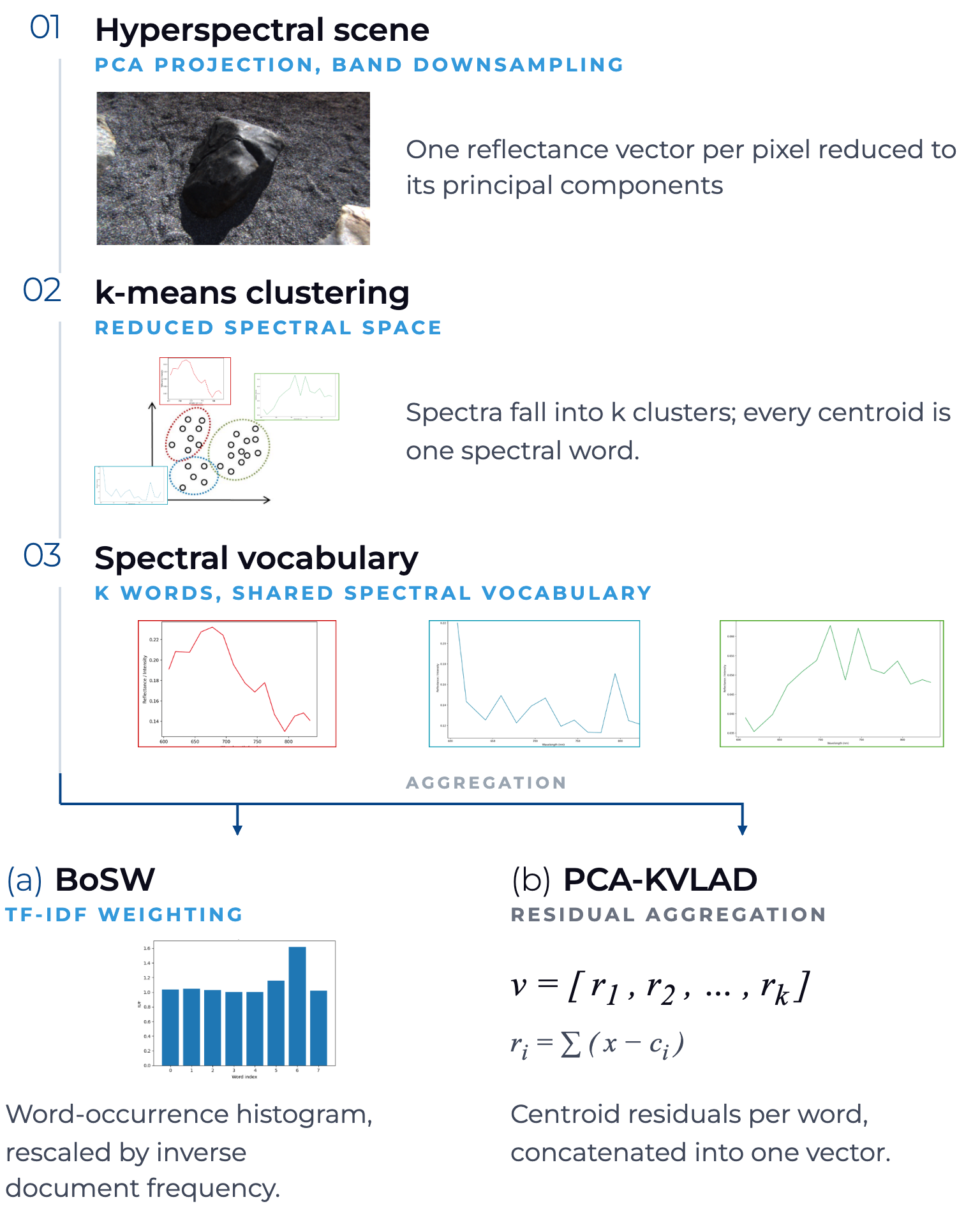}
  \caption{HSI descriptor extraction pipelines.
  (a)~BoSW constructs a TF-IDF weighted spectral word histogram;
  (b)~PCAK-VLAD aggregates VLAD residuals in PCA-reduced spectral space.
  Both share the same preprocessing, PCA, and K-Means vocabulary.}
  \label{fig:hsi}
\end{figure}

The hyperspectral sensor produces data cubes of shape
$1020\times540\times15$ after a structured preprocessing pipeline.
A descriptor pipeline is then applied, as shown in Fig.~\ref{fig:hsi}.

\subsubsection{Shared Preprocessing}
\begin{figure}[t]
  \centering
  \includegraphics[width=\columnwidth]{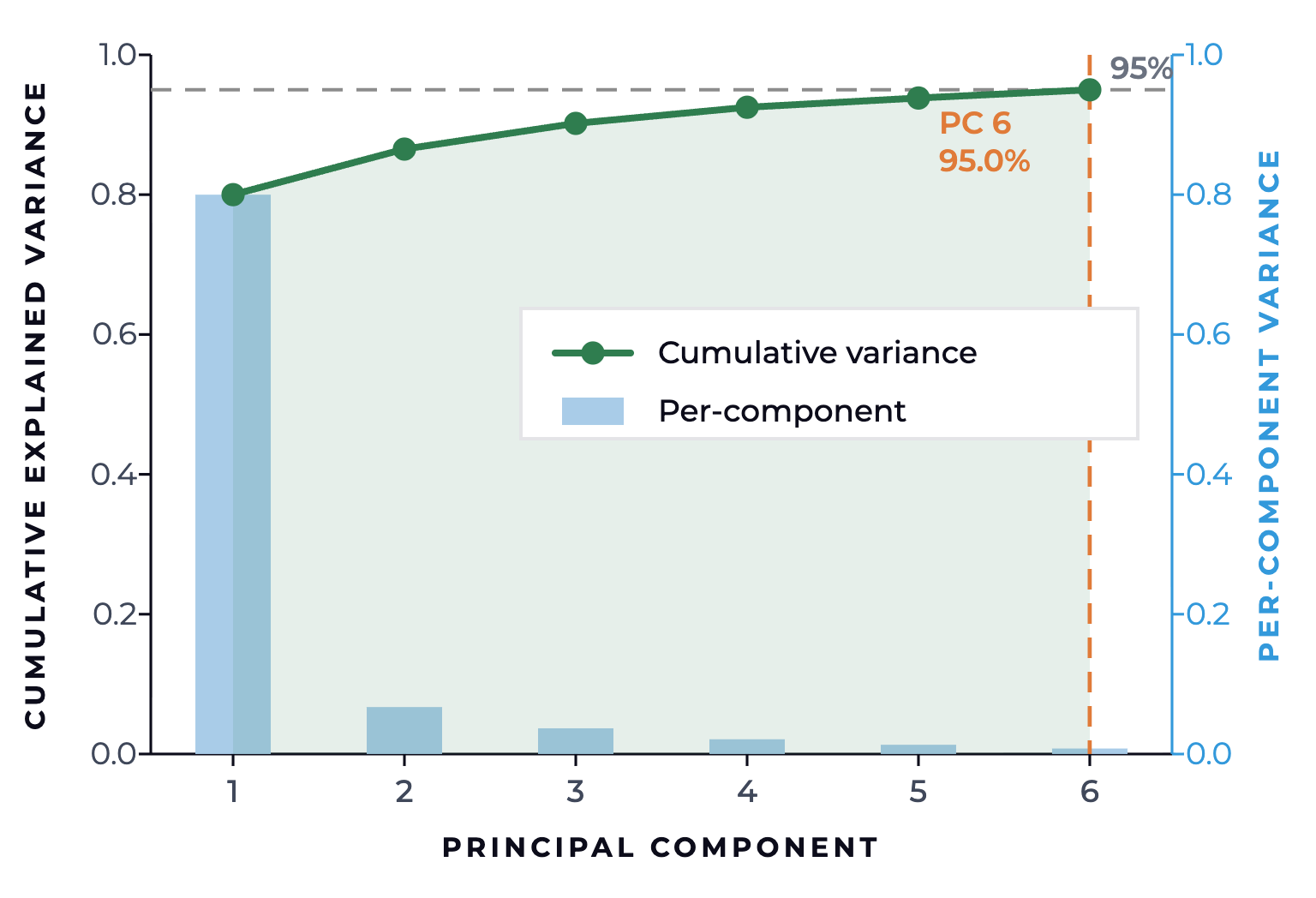}
  \caption{Cumulative explained variance of the fitted PCA model on
  the spectral training set. Six principal components capture
  $\approx$95\% of variance, justifying the $\mathbb{R}^{15}\!\to\!\mathbb{R}^{6}$
  dimensionality reduction. The steep drop after PC\,1 (80\% variance)
  reflects the dominant residual spectral shape mode; after per-pixel
  intensity normalisation (Eq.~\ref{eq:specnorm}), PC\,1 captures the
  largest source of remaining spectral variation across the training set,
  not raw reflectance intensity, which has already been divided out.}
  \label{fig:pca}
\end{figure}

Each cube is spatially downsampled by factor 4 via block averaging,
producing a $255\times135\times15$ tensor. Each pixel's spectral
signature $\mathbf{s}\in\mathbb{R}^{15}$ corresponds to the
calibrated reflectance vector $\mathbf{v}$ defined in Section~2.3 --
that is, the per-pixel 15-band reflectance after white/dark reference
correction and spectral crosstalk removal. This gives a physically
meaningful, sensor-corrected spectral measurement per pixel.
A further per-pixel spectral normalization is then applied:
\begin{equation}
  \mathbf{s}_{\mathrm{norm}} = \mathbf{s}\,/\,\mathrm{mean}(\mathbf{s}),
  \quad \mathrm{mean}(\mathbf{s}) = \tfrac{1}{15}\textstyle\sum_{b=1}^{15} s_b,
  \label{eq:specnorm}
\end{equation}
removing residual inter-frame intensity differences caused by illumination
variation or shadows while preserving the spectral shape that encodes
material identity.
Each cube is reshaped to an $N\times15$ matrix ($N = 255\times135 = 34{,}425$),
treating each pixel as an independent observation.
Principal Component Analysis (PCA) is then applied to reduce each
15-dimensional spectrum to $\mathbf{z}_i\in\mathbb{R}^{6}$, retaining
$\approx$95\% of spectral variance (Fig.~\ref{fig:pca}).

\subsubsection{Bag-of-Spectral-Words (BoSW)}
BoSW constructs a histogram-based descriptor encoding the distribution
of spectral material types per frame. For a query frame with $N'$ pixels:
(1) Each pixel spectrum is projected via PCA and assigned to its
nearest cluster: $a_i = \arg\min_k\|\mathbf{z}_i - \mathbf{c}_k\|$.
(2) Term frequency is computed: $\mathrm{tf}[k] = |\{i:a_i=k\}|/N'$.
(3) Inverse Document Frequency (IDF), learned from $M$ training frames,
re-weights by spectral word rarity:
\begin{equation}
  \mathrm{IDF}(k) = \log\!\left(\frac{M+1}{d_k+1}\right) + 1,
  \label{eq:idf}
\end{equation}
where $d_k$ is the number of training frames containing word $k$.
(4) Element-wise TF-IDF product and L2 normalisation yield
$\mathbf{d}_{\mathrm{BoSW}}\in\mathbb{R}^K$.
BoSW encodes spectral word frequency distributions (material type counts)
and is robust to noise under low spectral discriminability.

\subsubsection{PCA-K VLAD}
PCAK-VLAD replaces histogram counting with VLAD residual aggregation.
Using the same PCA and K-Means vocabulary, for each cluster $k$:
\begin{equation}
  \mathbf{v}_k = \sum_{i:\,a_i = k}\!\left(\mathbf{z}_i - \mathbf{c}_k\right) \in \mathbb{R}^6,
  \label{eq:specvlad}
\end{equation}
using the same nearest-neighbour assignment $a_i$ as in Eq.~\ref{eq:vlad}.
Residual vectors are concatenated, flattened to $\mathbb{R}^{6K}$,
and L2-normalised. PCAK-VLAD encodes intra-cluster residuals
(first-order moments relative to cluster centres), capturing both
spectral material distribution and intra-cluster variation, providing
greater expressiveness than BoSW but at higher sensitivity to noise.

\subsubsection{Computational Cost}
To quantify the computational cost of descriptor extraction, all
three pipelines were benchmarked on the full MMOTS-WS sequence, using
the same trained vocabularies employed in the final experiments
($K=8$ throughout). For each frame already resident in memory (sensor
read-out and disk I/O excluded), wall-clock time was measured for the
full compute pipeline using \texttt{time.perf\_counter()} after a
5-item warm-up on a single CPU thread (no GPU acceleration).
The benchmark processes each modality at its native captured frame rate
before synchronisation: 1{,}135 raw HSI frames and 1{,}506 raw RGB frames
(higher due to the RGB camera's faster capture rate), compared to the
1{,}072 synchronised RGB-HSI keyframe pairs reported in Table~\ref{tab:datasets}
and used in the SLAM evaluation.
On the HSI branch (spectral downsampling, radiometric normalisation,
PCA projection, K-Means cluster assignment, descriptor encoding), BoSW
required a mean of 14.09~ms per frame (median 13.70~ms, std 1.07~ms,
95th percentile 15.81~ms) and PCAK-VLAD required 14.22~ms (median
14.07~ms, std 0.62~ms, 95th percentile 15.14~ms) -- nearly identical,
as expected, since both share the same PCA-projection and
cluster-assignment cost and differ only in the final aggregation step.
On the RGB branch (grayscale conversion, SIFT detection and description,
K-Means assignment, VLAD aggregation), SIFT-VLAD required 12.60~ms per
frame (median 12.49~ms, std 1.03~ms, 95th percentile 13.66~ms). Since
each keyframe evaluates exactly one RGB descriptor together with one
HSI descriptor (Table~\ref{tab:exps}), the combined worst-case
descriptor overhead is $\approx$27~ms per keyframe -- well within the
sensor's 125~ms (8~Hz) capture interval, though this accounts only for
descriptor extraction and excludes the tracking, mapping, and fusion
stages of the full pipeline.

\begin{figure}[t]
  \centering
  \includegraphics[width=0.85\columnwidth]{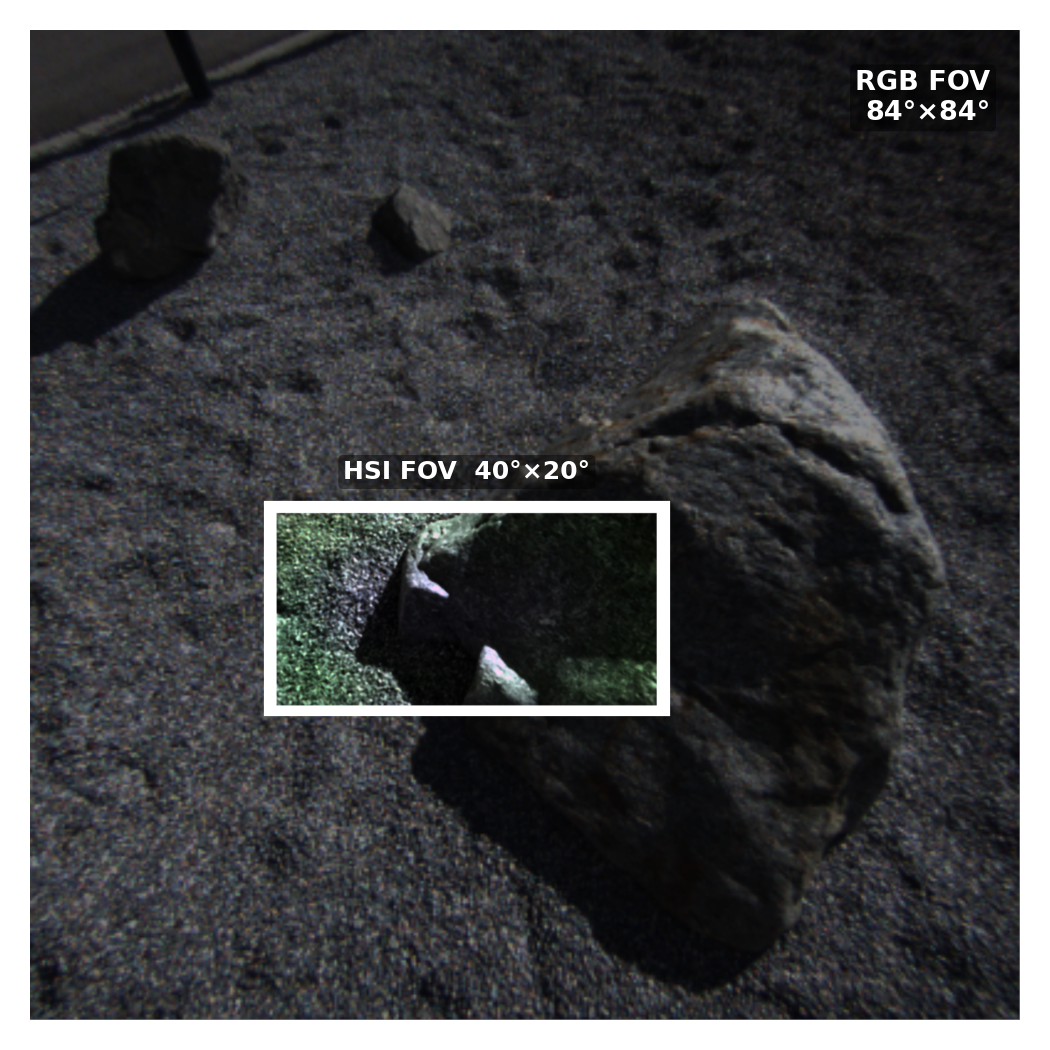}
  \caption{Field-of-view mismatch between the RGB camera
  (84$^\circ$$\times$84$^\circ$, full frame) and the Ximea IMEC
  hyperspectral sensor (40$^\circ$$\times$20$^\circ$, white rectangle).
  The HSI sensor covers only a central sub-region of the RGB frame,
  motivating decision-level fusion rather than pixel-level registration.}
  \label{fig:fov}
\end{figure}

\subsection{Decision-Level Fusion}

Decision-level fusion is adopted because the mismatched fields of
view between the RGB ($\approx$84$^\circ$$\times$84$^\circ$) and
HSI ($\approx$40$^\circ$$\times$20$^\circ$) cameras motivate
operating on per-frame descriptors rather than pixel-level
correspondence: the HSI sensor covers only a central sub-region of
the RGB frame (Fig.~\ref{fig:fov}), and reliable pixel-level
projection would require precise extrinsic calibration not available
in this setup.
Independent cosine similarity scores $S_{\mathrm{RGB}}$ and
$S_{\mathrm{HSI}}$ are therefore computed for each keyframe pair
and combined at the decision stage.

Two fusion strategies are evaluated:

\textbf{Strict-AND Fusion.}
A candidate is accepted only if both modalities independently
exceed their respective thresholds:
\begin{equation}
  S_{\mathrm{RGB}} \geq \tau_{\mathrm{RGB}}
  \;\;\text{AND}\;\;
  S_{\mathrm{HSI}} \geq \tau_{\mathrm{HSI}}.
  \label{eq:strict}
\end{equation}

\textbf{Weighted Fusion.}
A combined score is computed:
\begin{equation}
  S_{\mathrm{fused}} = \beta\cdot S_{\mathrm{RGB}} + (1-\beta)\cdot S_{\mathrm{HSI}},
  \label{eq:weighted}
\end{equation}
and accepted when $S_{\mathrm{fused}} \geq \tau_{\mathrm{fused}}$,
where $\tau_{\mathrm{fused}} = \beta\tau_{\mathrm{RGB}} + (1-\beta)\tau_{\mathrm{HSI}}$.
$\beta = 0.4$ is fixed prior to evaluation and not tuned on the test set
(i.e.\ HSI receives 60\% weight, based on the prior expectation that
spectral signatures would provide stronger discrimination than RGB on
featureless terrain; the experimental results do not confirm this prior).
Both $\tau_{\mathrm{RGB}}$ and $\tau_{\mathrm{HSI}}$ are set adaptively at runtime: for each query frame, $\tau$ is the minimum cosine similarity among the top-30\% most similar within-submap frame pairs under the respective descriptor, providing a data-driven threshold calibrated to local scene structure without using inter-submap ground-truth labels.

\section{Experimental Setup}

\subsection{Dataset and Test Site}

\begin{figure}[t]
  \centering
  \includegraphics[width=\columnwidth]{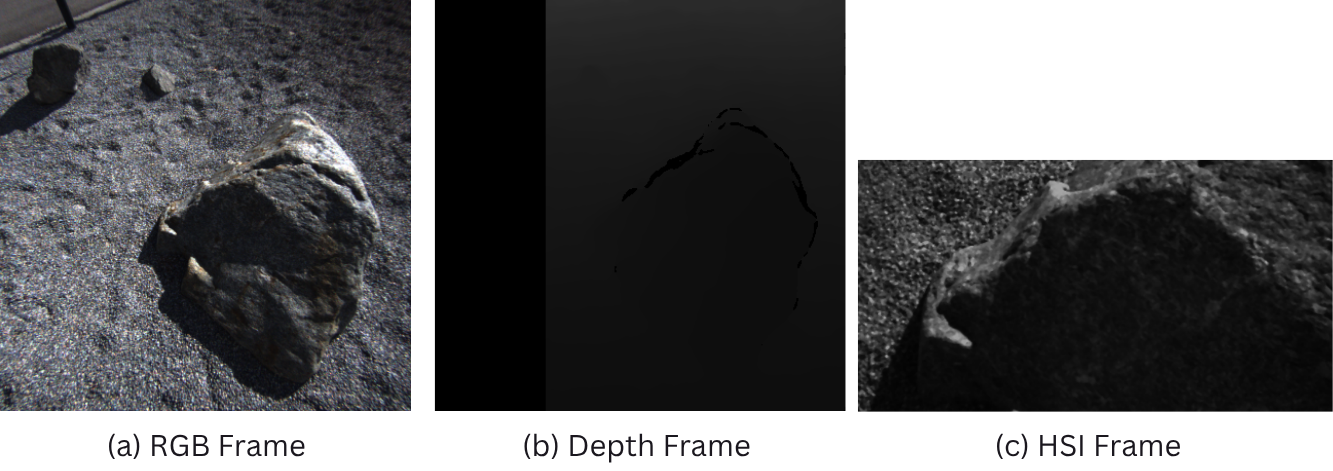}
  \caption{Synchronised multi-modal data sample from MMOTS-WS.
  (a)~RGB frame; (b)~Depth frame; (c)~HSI false-colour frame.
  Note the HSI frame covers a smaller field of view than the RGB pair.}
  \label{fig:data}
\end{figure}

Data was collected at the \textbf{Moon-Mars Outdoor Test Site (MMOTS)}
of DLR, Oberpfaffenhofen~\cite{dlrmmots}, a purpose-built planetary
analog environment with regolith-like soil, scattered rocks, gravel,
and craters simulating lunar and Martian surface conditions
(Fig.~\ref{fig:data}).

The Ximea IMEC hyperspectral camera~\cite{ximea2025} (Table~\ref{tab:sensor}) and a
stereo RGB pair were mounted on an Ardea multicopter~\cite{ardea2020}, operated
at low altitude above the analog terrain to simulate a rover-perspective sensor
view; the multicopter serves as a data-collection proxy and is not a planetary
rover itself.
RGB frames are rectified to $400\times400$~px (
FOV $\approx$84$^\circ$$\times$84$^\circ$). Depth maps are derived
via stereo disparity estimation. A hardware trigger synchronises RGB and HSI frame captures. Ground-truth poses are generated
using VINS-Fusion~\cite{qinlocal2019,qinglobal2019} in stereo mode with loop fusion enabled.
\begin{table}[t]
\centering
\caption{Ximea IMEC Hyperspectral Camera Specifications}
\label{tab:sensor}
\small
\begin{tabularx}{\columnwidth}{lX}
\hline
\textbf{Property} & \textbf{Value} \\
\hline
Spectral bands    & 15 channels \\
Spectral range    & 600--860~nm (Red-NIR) \\
Sensor resolution & 2048$\times$1088~px (native) \\
Cube resolution   & 1020$\times$540$\times$15 (after mosaic de-interleaving) \\
Frame rate        & up to 170~FPS (raw); operated at 8~Hz for HSI cubes \\
Pixel size        & 5.5~$\mu$m \\
Shutter           & Global \\
Focal Length      & 16~mm \\
FOV               & $\approx$40$^\circ$$\times$20$^\circ$ \\
Interface         & USB3 (5~Gbit/s) \\
\hline
\end{tabularx}
\end{table}

\begin{figure}[t]
  \centering
  \includegraphics[width=\columnwidth]{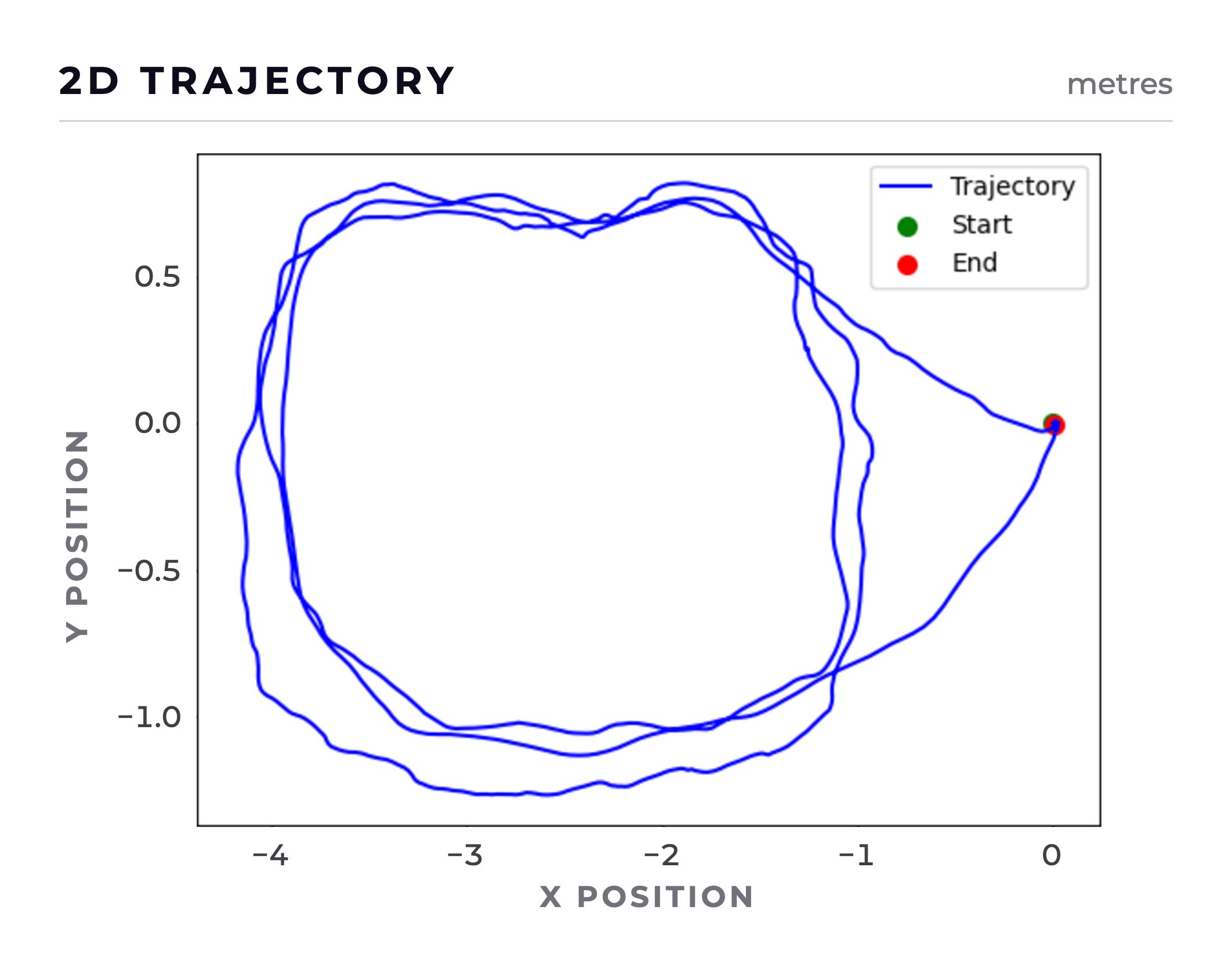}
  \caption{Ground-truth trajectory of the MMOTS-WS dataset.
  The loop closure is enforced by starting and ending at the
  same spatial location.}
  \label{fig:trajectory}
\end{figure}

Table~\ref{tab:datasets} summarises all collected sequences.
\textbf{MMOTS-WS} is the primary evaluation sequence: three loop traversals
over an 8~m-extent area, all beginning and ending at the same spatial
location, providing ground-truth loop pairs for quantitative assessment
(Fig.~\ref{fig:trajectory}).
The three additional MMOTS trajectories---MMOTS-LL, MMOTS-RL, and MMOTS-US---are
longer traversals (15--24.5~m) recorded at the same site under varied walking
patterns; they do not share MMOTS-WS's closed-loop geometry and therefore yield
no ground-truth loop pairs suitable for trajectory evaluation.
SS-SWEDEN is an outdoor sequence collected at a natural rocky terrain site in
Sweden, covering 123.3~m over 262.2~s; it offers spectral diversity from a
geographically distinct environment but was acquired without a pose reference
system, precluding trajectory evaluation.
All sequences are used jointly to train the spectral vocabulary: K-Means
codebook, PCA basis, and IDF weights. Because vocabulary learning is
unsupervised (K-Means, TF-IDF), including MMOTS-WS in training does not
constitute label leakage in the strict sense, though spectral vocabulary
statistics are mildly influenced by the MMOTS-WS frame distribution.
The codebook size $K=8$ is selected based on positive/negative cosine
similarity separation statistics computed on MMOTS-WS (Section~6.1),
which is also the sole evaluation sequence; this represents a form of
hyperparameter tuning on the test set and is acknowledged as a
limitation. All experiments use random seed 0.

\begin{table}[t]
\centering
\caption{Overview of Collected Datasets}
\label{tab:datasets}
\small
\begin{tabularx}{\columnwidth}{lXXXX}
\hline
\textbf{Dataset} & \textbf{RGB} & \textbf{HSI} & \textbf{Dur.(s)} & \textbf{Dist.(m)} \\
\hline
MMOTS-WS  & 1072  & 1072  & 133.9 & 8    \\
MMOTS-LL  & 1385  & 1385  & 172.9 & 15   \\
MMOTS-RL  & 945   & 945   & 119.1 & 18   \\
MMOTS-US  & 1540  & 1540  & 186.0 & 24.5 \\
SS-SWEDEN & 2100  & 2100  & 262.2 & 123.3 \\
\hline
\textbf{Total} & \textbf{7042} & \textbf{7042} & \textbf{874.1} & \textbf{188.8} \\
\hline
\end{tabularx}
\end{table}

\subsection{Experiment Configurations}

Ten experiments are conducted on MMOTS-WS, building progressively
from baseline to full multi-modal system (Table~\ref{tab:exps}).
E1 applies the unmodified LoopSplat configuration to establish a
failure-mode baseline. E2 introduces outdoor-specific parameter tuning
with NetVLAD, establishing the RGB-only reference. E3 replaces NetVLAD
with SIFT-VLAD. Experiments E4 and E5 introduce HSI under Strict-AND
and Weighted fusion respectively, across all descriptor combinations.

\begin{table}[t]
\centering
\caption{Experiment Configuration Matrix}
\label{tab:exps}
\small
\begin{tabularx}{\columnwidth}{lXXX}
\hline
\textbf{ID} & \textbf{RGB} & \textbf{HSI} & \textbf{Fusion} \\
\hline
E1  & NetVLAD (default) & --       & -- \\
E2  & NetVLAD (tuned)   & --       & -- \\
E3  & SIFT-VLAD K=8     & --       & -- \\
E4-A & NetVLAD         & BoSW      & Strict-AND \\
E4-B & SIFT-VLAD       & BoSW      & Strict-AND \\
E4-C & NetVLAD         & PCAK-VLAD & Strict-AND \\
E4-D & SIFT-VLAD       & PCAK-VLAD & Strict-AND \\
E5-A & NetVLAD         & BoSW      & Weighted \\
E5-B & SIFT-VLAD       & BoSW      & Weighted \\
E5-C & NetVLAD         & PCAK-VLAD & Weighted \\
E5-D & SIFT-VLAD       & PCAK-VLAD & Weighted \\
\hline
\end{tabularx}
\end{table}

\subsection{Evaluation Metrics}

\textbf{Trajectory accuracy} is measured by the Absolute Trajectory
Error (ATE) RMSE after a similarity transformation (Sim(3)) alignment:
\begin{equation}
  \mathrm{ATE}_{\mathrm{RMSE}} =
  \sqrt{\frac{1}{N}\sum_{i=1}^{N}\|\mathbf{x}_{\mathrm{est},i}
  - \mathbf{x}_{\mathrm{gt},i}\|^2}.
  \label{eq:ate}
\end{equation}
The Sim(3)-aligned variant removes global reference frame offset and
scale and is the primary reported metric. For a metric stereo/RGB-D
system, scale is nominally observable from depth; Sim(3) alignment
may therefore absorb residual scale error and report optimistically low
ATE values. SE(3)-aligned ATE (no scale correction) was not computed
and represents an acknowledged limitation of the current evaluation;
relative comparisons between configurations are nonetheless valid
since all are evaluated under the same alignment.
\textbf{Loop closure detection} is evaluated via precision,
recall, and F1-score against a set of ground-truth loop pairs.
A pair $(i,j)$ qualifies as a ground-truth loop closure if the two frames
lie within 0.2~m translation and 1$^\circ$ rotation in ground-truth pose,
with a minimum separation of 50 frames and 10~s to exclude temporally
adjacent observations.
$\mathcal{D}$ is the set of detected closures and
$\mathcal{G}$ is the set of ground-truth closures:
\begin{align}
  \mathrm{Prec.} &= |\mathcal{D}\cap\mathcal{G}|/|\mathcal{D}|, \quad
  \mathrm{Rec.}  = |\mathcal{D}\cap\mathcal{G}|/|\mathcal{G}|, \nonumber\\
 \mathrm{F1}    &= 2\cdot\frac{\mathrm{Prec.}\times\mathrm{Rec.}}{\mathrm{Prec.}+\mathrm{Rec.}}.
  \label{eq:prf}
\end{align}

\section{Descriptor Analysis}

\subsection{RGB Descriptor Analysis}

Table~\ref{tab:rgbcosine} reports cosine similarity statistics for
positive (ground-truth loop closure) and negative (non-matching) pairs
across all RGB descriptor configurations on the full MMOTS-WS dataset.

\begin{table}[t]
\centering
\caption{Cosine Similarity Statistics for RGB Descriptors (MMOTS-WS)}
\label{tab:rgbcosine}
\small
\begin{tabularx}{\columnwidth}{lXXXX}
\hline
\textbf{Descriptor} & \textbf{Pos.}$\,\mu$ & \textbf{Pos.}$\,\sigma$ & \textbf{Neg.}$\,\mu$ & \textbf{Neg.}$\,\sigma$ \\
\hline
NetVLAD        & 0.662 & 0.137 & 0.247 & 0.137 \\
SIFT-VLAD K=8  & 0.729 & 0.105 & 0.404 & 0.221 \\
SIFT-VLAD K=16 & 0.581 & 0.125 & 0.224 & 0.139 \\
SIFT-VLAD K=32 & 0.446 & 0.123 & 0.144 & 0.105 \\
SIFT-VLAD K=64 & 0.347 & 0.110 & 0.087 & 0.090 \\
\hline
\multicolumn{5}{l}{\small NV = NetVLAD, SV = SIFT-VLAD}
\end{tabularx}
\end{table}

The pretrained NetVLAD descriptor shows a distinctive mean separation
between positive and negative pairs. However, both distributions contain
outliers, producing overlap between true and false matches. This issue
is largely attributed to the domain gap between the data used to train
NetVLAD's underlying VGG-16 backbone, which favours object-centric urban
scenes, and the MMOTS-WS dataset, which primarily features unstructured
outdoor terrain of stones and gravel.

SIFT-VLAD with K=8 shows positive pairs tightly grouped around a high
mean, with minimal overlap against the negative distribution. This
indicates robustness for loop closure detection at this setting.
As K increases, the positive-pair mean decreases while separation
between positives and negatives slightly improves. This trade-off
between lower absolute similarity for true matches and better
separability means that higher K values are less practical for
reliable retrieval. Additionally, repetitive textures such as gravel
and stone lead to recurring SIFT keypoints across different locations,
which can limit discriminative power when relying solely on SIFT features.

Overall, SIFT-VLAD with K=8 emerges as the best configuration for this
dataset, balancing cluster quality, compact positive-pair distributions,
and sufficient separation from negatives.

\begin{figure}[t]
  \centering
  \includegraphics[width=\columnwidth]{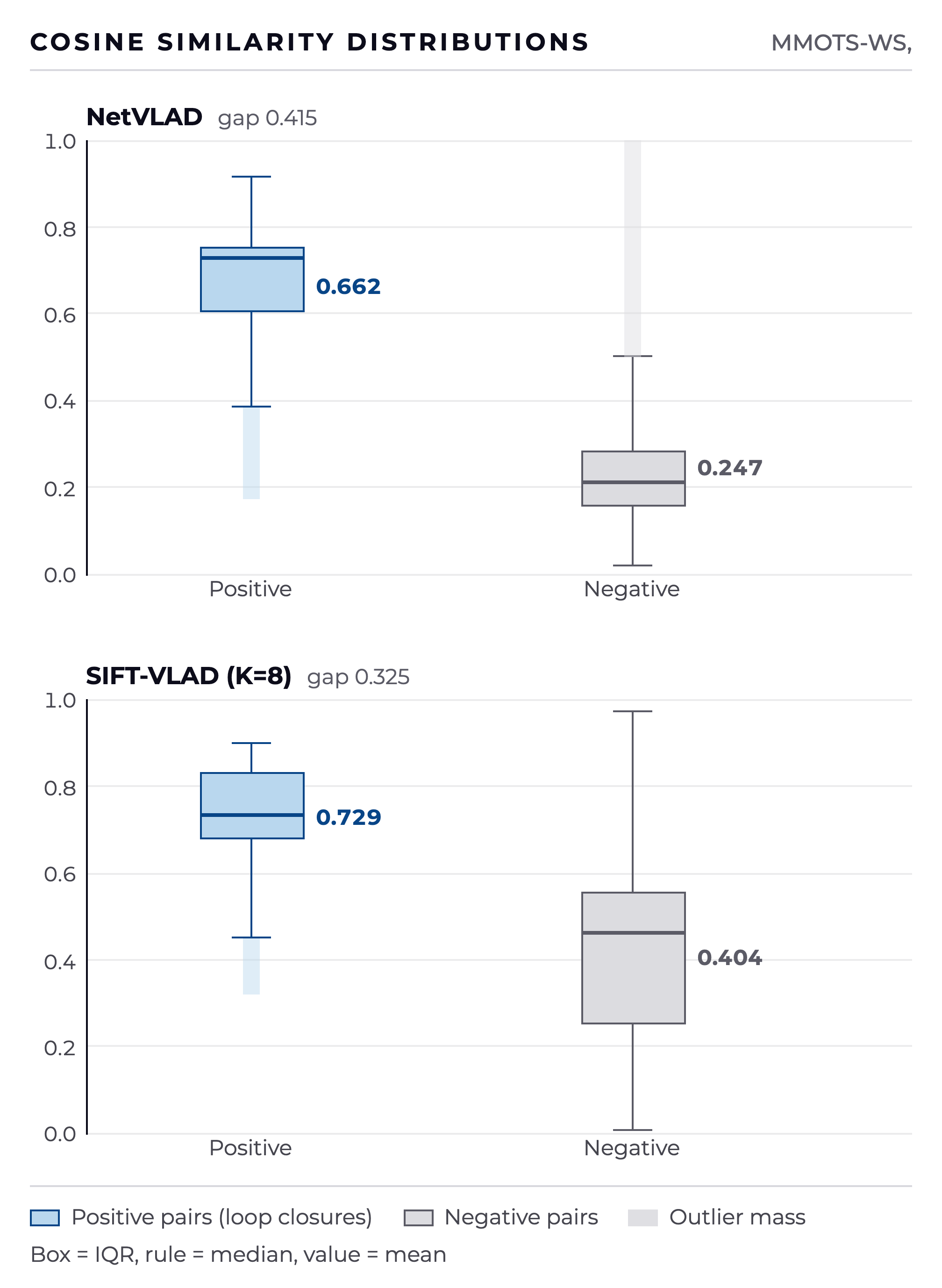}
  \caption{Cosine similarity distributions (positive and negative pairs)
  for SIFT-VLAD K=8 and NetVLAD on the full MMOTS-WS dataset.}
  \label{fig:boxplots}
\end{figure}

\subsection{HSI Descriptor Analysis}

\textbf{Cluster quality.}
The quality of the clustering process was assessed by varying the number
of clusters (K=8, 16, 32, 64) and analysing the centroid spectra.
A smaller K yields clearer cluster separations and higher
silhouette scores (0.232 at K=8, decreasing to 0.178 at K=16 and
0.152 at K=32), indicating more meaningful spectral groupings at low K.
Increasing K captures finer-grained distinctions but leads to less
compact and more fragmented clusters. Silhouette scores remain
relatively low across all values of K, suggesting there is scope for
improvement in the spectral preprocessing pipeline.

\begin{figure}[t]
  \centering
  \includegraphics[width=0.9\columnwidth]{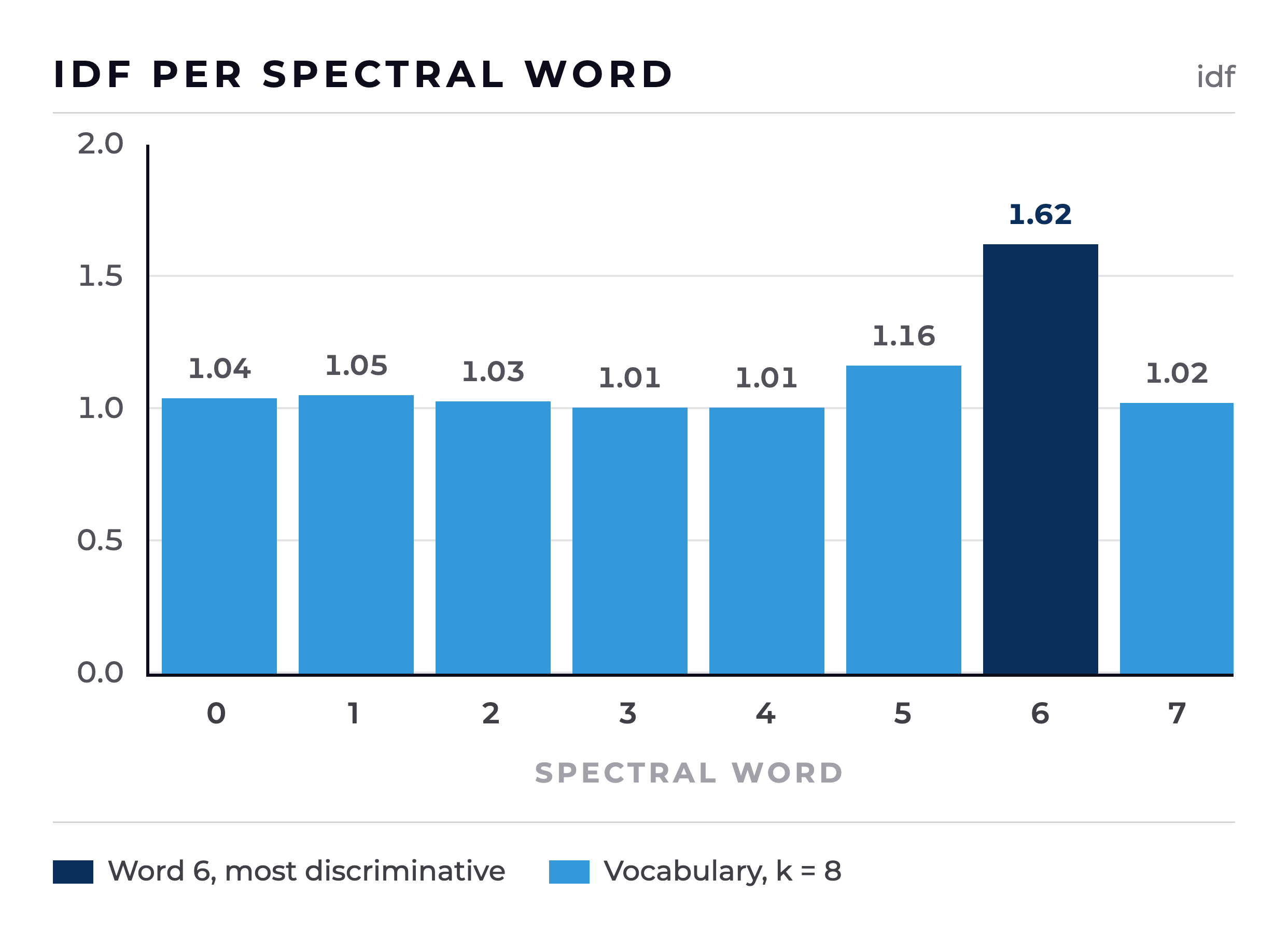}\\[2pt]
  \includegraphics[width=0.9\columnwidth]{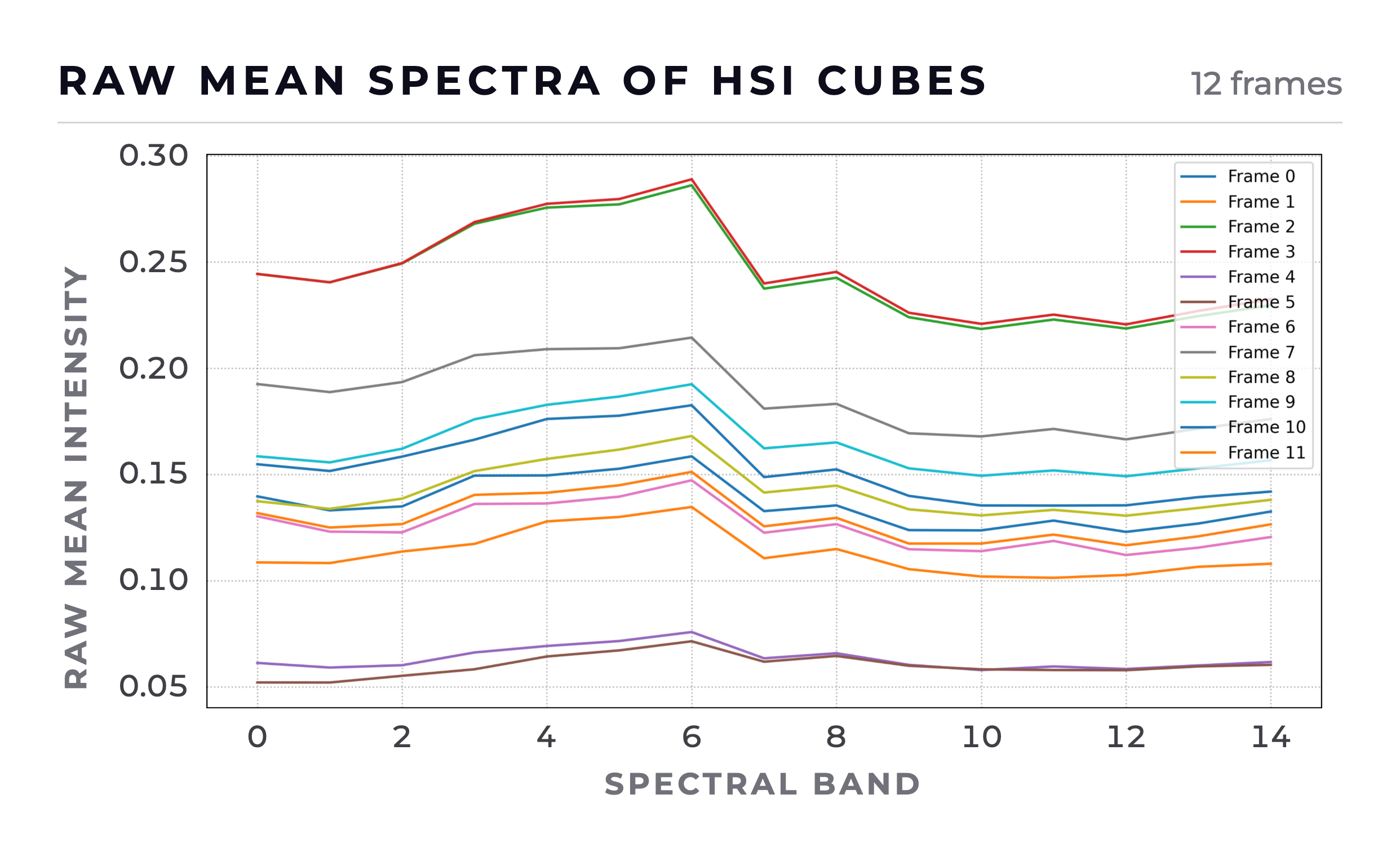}
  \caption{Top: IDF weights per spectral word (K=8); spectral word~6
  receives the highest IDF weight ($\approx$1.6), identifying a rare material.
  Bottom: Raw mean spectra of all MMOTS-WS HSI keyframe cubes.
  Curves are largely similar in shape, differing mainly in intensity,
  indicating limited material discriminability in the 600--860~nm range.}
  \label{fig:spectra}
\end{figure}

\textbf{IDF weighting.}
The BoSW model applies Inverse Document Frequency (IDF) weighting
learned from the full dataset. Most spectral words receive similar
IDF weights, indicating limited spectral diversity across the captured
frames. However, spectral word~6 receives a distinctly higher IDF
weight ($\approx$1.6), indicating that the BoSW model assigned
elevated discriminative weight to a spectrally rare pattern in the
training data. This confirms that the model can identify and up-weight
rare spectral signatures when they appear.

\textbf{Raw mean spectral signatures.}
Fig.~\ref{fig:spectra} shows the raw mean spectral signatures of all
HSI keyframe cubes from the MMOTS-WS dataset. The spectral curves
appear largely similar across frames, differing mainly in absolute
intensity. These intensity differences may result from varying rock
reflectivity, or the presence of shadows. The similarity in curve shape indicates that the sensor's
spectral range (600--860~nm) may not be sufficient to discriminate
between the materials present at this outdoor test site. This finding
directly explains why HSI fusion does not consistently outperform the
RGB-only baseline and motivates the use of SWIR-capable sensors in
future work.

\section{Results}

\subsection{Trajectory Accuracy}

\begin{figure}[t]
  \centering
  \includegraphics[width=\columnwidth]{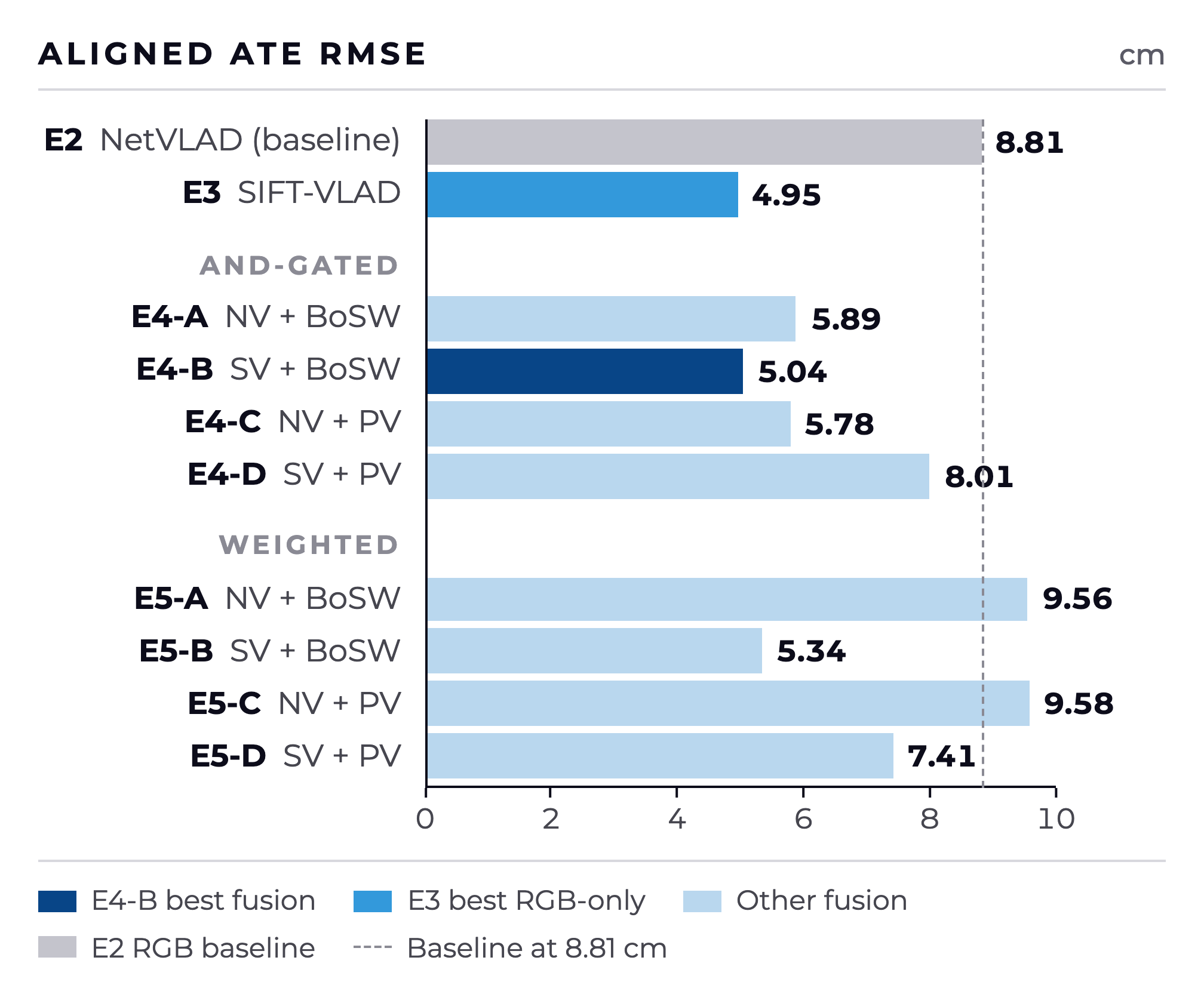}
  \caption{Aligned ATE RMSE for tuned experiments E2--E5-D on MMOTS-WS.
  E3 (SIFT-VLAD, RGB-only) achieves the best overall result (4.95~cm);
  E4-B (SIFT-VLAD + BoSW, Strict-AND) achieves the best fusion result
  (5.04~cm), within 1.8\% of E3. The gap to E2 (8.81~cm) is attributable
  to the RGB descriptor change (NetVLAD~$\to$~SIFT-VLAD), not to HSI fusion.
  NV: NetVLAD; SV: SIFT-VLAD; PV: PCAK-VLAD; AND: Strict-AND; Wt.: Weighted.}
  \label{fig:ate_bar}
\end{figure}

The default LoopSplat configuration fails catastrophically on
outdoor data (3380.89~cm aligned ATE), confirming that
the indoor-tuned pipeline cannot be directly applied to planetary
analog terrain. Fig.~\ref{fig:ate_bar} shows
ATE RMSE results for all tuned experiments (E2--E5-D) on MMOTS-WS.
With outdoor-specific tuning (E2), the system achieves
8.81~cm. Replacing NetVLAD with SIFT-VLAD (E3) improves this to
4.95~cm, validating the domain-mismatch hypothesis.

Introducing HSI under Strict-AND fusion preserves trajectory accuracy
relative to the best RGB-only configuration. The best fusion result,
E4-B (SIFT-VLAD + BoSW, Strict-AND), achieves 5.04~cm aligned ATE
RMSE, statistically comparable to the RGB-only counterpart E3
(4.95~cm, a 1.8\% difference). The controlled comparison for assessing
the HSI contribution is E4-B versus E3, where the RGB backbone is held
fixed at SIFT-VLAD; under this comparison, HSI fusion results in a
marginal 1.8\% increase in ATE error (4.95~cm to 5.04~cm). The 43\% improvement relative to
E2 (8.81~cm) is attributable to the RGB descriptor change from NetVLAD
to SIFT-VLAD: the descriptor change alone (E2~$\to$~E3) already yields
43.8\% improvement without any HSI component. E3
(SIFT-VLAD, RGB-only) achieves the best overall trajectory accuracy at
4.95~cm, with the sensor's limited spectral range identified as the
primary limiting factor (Section~8). The PCAK-VLAD variant
(E4-D) performs worse (8.01~cm), consistent with the spectral analysis
finding. Under Weighted fusion, E5-B (SIFT-VLAD + BoSW) achieves
5.34~cm, close to E4-B, but other weighted combinations degrade
substantially, confirming that unconstrained HSI score weighting
contaminates reliable RGB decisions when spectral discriminability is low.

\subsection{Loop Closure Detection}

\begin{figure}[t]
  \centering
  \includegraphics[width=\columnwidth]{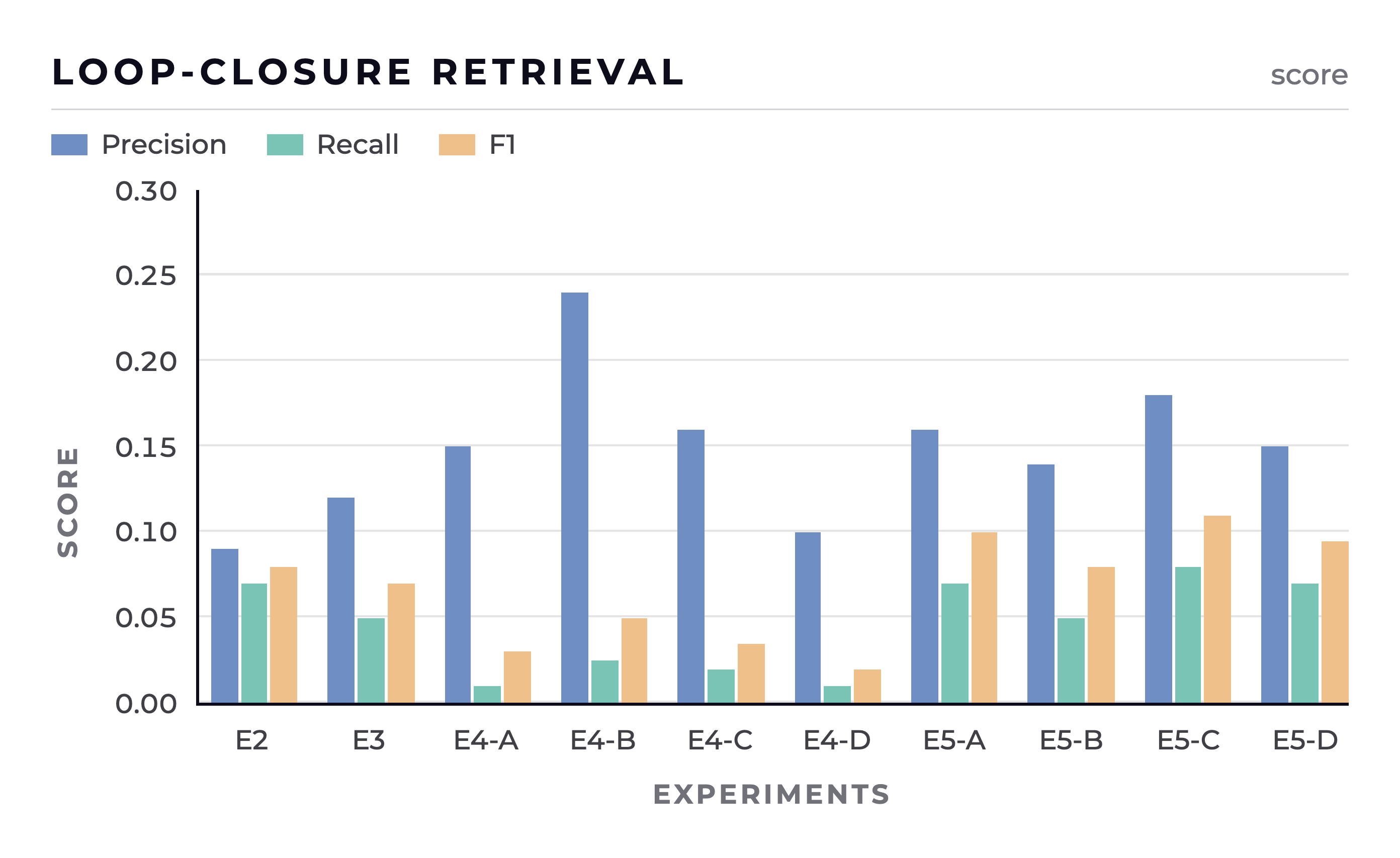}
  \caption{Precision, Recall, and F1-score across all experiments,
  computed before geometric verification (descriptor-level retrieval only).
  E4-B achieves the highest precision ($\approx$0.24); E5-C achieves the
  best F1-score balance ($\approx$0.11).
  NV: NetVLAD; SV: SIFT-VLAD; PV: PCAK-VLAD; AND: Strict-AND; Wt.: Weighted.}
  \label{fig:lc}
\end{figure}

As shown in Fig.~\ref{fig:lc}, precision consistently exceeds
recall across all experiments. These metrics are computed before
geometric verification (Gaussian-overlap check), so the reported
precision and recall reflect the descriptor-level retrieval stage only;
post-verification loop closure counts are not reported separately,
as the geometric check is embedded within LoopSplat's submap registration
and its output is not exposed as a standalone metric in the current
evaluation pipeline.
E4-B achieves the highest precision ($\approx$0.24), meaning
approximately 76\% of detections are false positives under the stated
evaluation criteria; recall and F1 remain very low across all
configurations. E5-C (NetVLAD + PCAK-VLAD, Weighted) achieves the
best F1 balance ($\approx$0.11) by trading some precision for improved
recall. These results indicate that no configuration achieves
satisfactory retrieval performance in absolute terms; fusion-based
approaches modestly reduce false positives relative to RGB-only
retrieval under the Strict-AND gate, but the overall detection rate
remains low. The uniformly low recall reflects both the limited number
of successfully detected loop closures and the inherent difficulty
of the dataset: spectrally similar terrain reduces HSI discriminability,
while perceptual aliasing limits RGB retrieval.

\section{Discussion}

\textbf{RGB descriptor performance.}
Across all experiments, SIFT-VLAD consistently outperforms NetVLAD on the MMOTS-WS outdoor dataset.
This is expected, as NetVLAD is trained on urban, object-centric imagery and struggles in scenes dominated
by uniform textures such as gravel and sand~\cite{netvlad2016,altawil2024}. SIFT, by contrast, captures
fine-grained local features such as edges and small stones, making it better suited to outdoor planetary analog terrain.

\textbf{HSI descriptor performance.}
Among the two HSI descriptor models, BoSW consistently outperforms PCAK-VLAD. The sensor's spectral
range (600--860~nm) produces spectra that are highly similar across different terrain patches, leaving
little inter-frame residual variation for PCAK-VLAD to encode. The simpler frequency histogram of BoSW
is more robust under these low-discriminability conditions.

\textbf{Fusion strategy.}
Strict-AND fusion yields better results than Weighted fusion in most configurations. By requiring both
RGB and HSI descriptors to independently agree on a loop closure candidate, Strict-AND avoids introducing
false positives when HSI discriminability is low. Weighted fusion, by contrast, allows uncertain HSI scores
to influence the decision, which degrades performance when the spectral signal is weak.

\textbf{Sensor limitations and the path to planetary SLAM.}
A central finding of this work is that the Red-NIR spectral range (600--860~nm) provides limited
discriminability for planetary analog terrain composed mainly of repetitive materials such as rocks and gravel.
The manufacturer originally designed this camera for agricultural vegetation analysis; its spectral range
yields little material-specific information in this planetary setting.
The SWIR range (900--2500~nm) captures characteristic absorption features of common planetary minerals,
where reflectance dips at specific wavelengths uniquely fingerprint mineral composition~\cite{murchie2007,lee2021}.
On the \textbf{Moon}, the two dominant terrains are spectrally separable only in SWIR:
highland anorthosite (plagioclase, $\sim$1250~nm) is distinct from mare basalt
(high-Ca pyroxene Band~I at 1000--1050~nm; Band~II at 2200--2300~nm),
while olivine produces a broad asymmetric absorption centred at 1040--1090~nm~\cite{lucey2006}.
On \textbf{Mars}, the three major mineral eras~\cite{murchie2007,bibring2006} each have diagnostic SWIR signatures:
ancient phyllosilicates (smectite clays, 2200--2350~nm; Fe-OH at 2290~nm),
Hesperian sulfates (gypsum at 2220~nm; jarosite at 2270~nm),
and mafic pyroxenes (Band~I 900--1050~nm; Band~II 1800--2300~nm).
None of these features fall within the 600--860~nm range of the current sensor.
This is the primary hypothesis for why Red-NIR HSI provides limited advantage
over RGB for planetary rock terrain; however, the experiments establish this for
the specific sensor, preprocessing pipeline, descriptor family, and test site
used here, and cannot isolate wavelength range as the sole cause. Other
contributing factors include descriptor design, FOV mismatch, spatial
aggregation, and material homogeneity at the MMOTS test site.
Controlled SWIR measurements or labeled material spectra would be required to
confirm the spectral range hypothesis directly.
This is consistent with orbital instruments (CRISM~\cite{murchie2007}, M$^3$~\cite{pieters2009}) that operate in SWIR and have successfully
mapped mineral diversity on Mars and the Moon.
However, even SWIR spectroscopy faces a fundamental planetary challenge: space weathering deposits
nanophase iron on exposed surfaces, suppressing spectral contrast and reducing discrimination
between rock types~\cite{hapke2001}.
This motivates a broader multi-modal approach for robust planetary SLAM:
RGB captures visual texture, SWIR-HSI encodes material identity, and LiDAR provides
geometric distinctiveness through rock shape and relief--three complementary channels
that are individually limited but jointly robust.
Crucially, HyperLoop's decision-level fusion architecture accommodates additional modalities
without structural change: a LiDAR-derived geometric descriptor can be fused alongside
RGB and HSI scores using the same Strict-AND or Weighted gate, making the framework
directly extensible to the full sensor suite required for planetary exploration.

\textbf{System and hardware limitations.}
The hyperspectral camera's narrow field of view ($\approx$40$^\circ$$\times$20$^\circ$) compared to the
RGB camera ($\approx$84$^\circ$$\times$84$^\circ$) reduces spatial coverage per frame and complicates
alignment between modalities. Some hyperspectral channels also suffer from low resolution and blurriness,
affecting descriptor quality. Additionally, operating RGB and HSI sensors under ROS~1 and ROS~2
respectively introduced synchronisation challenges, with occasional frame drops compensated by
nearest-timestamp substitution.

\textbf{LoopSplat baseline limitations.}
LoopSplat was originally designed and tuned for structured indoor environments, and required significant
hyperparameter adjustments to operate on the outdoor MMOTS-WS dataset. LoopSplat's geometric verification relies on Gaussian splat overlap alone, without accounting for pose
uncertainty; this is an inherited limitation that can cause valid loop closures to be rejected under
accumulated drift. Pose graph optimisation is only triggered when new loop edges are added.

\textbf{Dataset constraints.}
The evaluation is limited to a single planetary analog site with a relatively small number of traverses,
and trajectory evaluation is conducted on a single 8~m closed-loop sequence.
This restricts the diversity of conditions tested and precludes claims about
generalisation to other sites, illumination conditions, or terrain types.
Additionally, the codebook size $K=8$ was selected based on positive/negative
similarity statistics computed on MMOTS-WS, which is also the sole evaluation
sequence; results at this setting may therefore carry a degree of
test-set optimism. A leave-one-sequence-out or leave-one-site-out protocol
would be required for unbiased hyperparameter selection.

\section{Conclusions}

This paper presents HyperLoop, the first framework integrating
hyperspectral imaging into the loop closure module of a 3DGS-based
dense SLAM system. The central empirical finding is that the Red-NIR spectral range
(600--860~nm) is insufficient for discriminating planetary analog terrain.
The best fusion configuration (SIFT-VLAD + BoSW, Strict-AND, 5.04~cm)
is within 1.8\% of the RGB-only SIFT-VLAD baseline (4.95~cm) and does
not surpass it. The 43\% reduction in aligned ATE RMSE over the
NetVLAD baseline (8.81~cm) is attributable entirely to replacing
NetVLAD with SIFT-VLAD, not to HSI fusion.
This null result has direct mission implications: SWIR-capable
sensors (900--2500~nm), which capture diagnostic absorption features
of pyroxene, olivine, and feldspar, are required to fully exploit
material semantics for autonomous rover SLAM.

Future directions include:
\begin{enumerate}[leftmargin=1.5em, noitemsep, topsep=2pt]
  \item Upgrading to a SWIR-capable sensor (900--2500~nm) for rock
        mineral discrimination.
  \item Developing end-to-end learned descriptors that jointly encode
        spectral and visual appearance for place recognition.
  \item Extending HSI integration into tracking, mapping, and
        PGO information matrix estimation.
  \item Evaluating on additional planetary analog sites (Atacama
        Desert, Iceland lava fields) to assess generalisation.
  \item Incorporating adaptive exposure control for the hyperspectral
        camera to mitigate intensity variation across frames under
        varying illumination conditions.
  \item Extending the decision-level fusion architecture to a full
        multi-modal suite (RGB~+~SWIR-HSI~+~LiDAR), exploiting
        geometric distinctiveness from LiDAR alongside spectral
        material identity from SWIR to address space weathering
        and perceptual aliasing jointly.
\end{enumerate}

\section*{Acknowledgements}

The authors thank Julius Quell for assistance with dataset collection and ground-truth generation using VINS-Fusion.

\clearpage
\bibliographystyle{unsrt}
\bibliography{biblio}

\end{document}